\documentclass{article}

\usepackage[preprint]{neurips_2025}

\usepackage[utf8]{inputenc} 
\usepackage[T1]{fontenc}    
\usepackage{hyperref}       
\usepackage{url}            
\usepackage{booktabs}       
\usepackage{amsfonts}       
\usepackage{nicefrac}       
\usepackage{microtype}      
\usepackage{xcolor}         

\usepackage{amsfonts}       
\usepackage{amsmath}
\usepackage{amssymb}
\usepackage{comment}

\title{Do It for HER: First-Order Temporal Logic Reward \\Specification in Reinforcement Learning \\ (Extended Version)}

\newif\ifshownotes
\usepackage[textsize=tiny]{todonotes}
\newcommand{\nb}[3]{\ifshownotes{\colorbox{#2}{\bfseries\sffamily\scriptsize\textcolor{white}{#1}}}{\textcolor{#2}{\sf\small\emph{#3}}}\fi}
\newcommand{\ale}[1]{\nb{Alessandro}{yellow!70!black}{#1}}
\newcommand{\mat}[1]{\nb{Matteo}{red}{#1}}
\newcommand{\pier}[1]{\nb{Pier}{blue}{#1}}
\newcommand{\fausto}[1]{\nb{Fausto}{green}{#1}}

\newcommand{\lnext}{X}
\newcommand{\lwnext}{\tilde{X}}

\newcommand{\lnextmt}{%
  \mathord{\text{\textcircled{$\scriptstyle\phantom{\sim}$}}}%
}

\newcommand{\lwnextmt}{%
  \mathord{\text{\textcircled{$\scriptstyle\sim$}}}%
}

\author{
  Pierriccardo Olivieri \\
  DEIB, Politecnico di Milano\\
  Milan, Italy \\
  \texttt{pierriccardo.olivieri@polimi.it} \\
  \And
  Fausto Lasca \\
  DEIB, Politecnico di Milano \\
  Milan, Italy  \\
  \texttt{fausto.lasca@mail.polimi.it} \\
  \And
  Alessandro Gianola \\
  INESC-ID/Instituto Superior Técnico\\
  Universidade de Lisboa \\
  Lisbon, Portugal \\
  \texttt{alessandro.gianola@tecnico.ulisboa.pt} \\
  \And
  Matteo Papini \\
  Dipartimento di Informatica \\
  Università degli Studi di Milano \\
  Milan, Italy \\
  \texttt{matteo.papini@unimi.it} \\
}

\begin{document}

\maketitle

\begin{abstract}
In this work, we propose a novel framework for the logical specification of non-Markovian rewards in Markov Decision Processes (MDPs) with large state spaces. Our approach leverages Linear Temporal Logic Modulo Theories over finite traces (LTLfMT), a more expressive extension of classical temporal logic in which predicates are first-order formulas of arbitrary first-order theories rather than simple Boolean variables. This enhanced expressiveness enables the specification of complex tasks over unstructured and heterogeneous data domains, promoting a unified and reusable framework that eliminates the need for manual predicate encoding. However, the increased expressive power of LTLfMT introduces additional theoretical and computational challenges compared to standard LTLf specifications. We address these challenges from a theoretical standpoint, identifying a fragment of LTLfMT that is tractable but sufficiently expressive for reward specification in an infinite-state-space context. From a practical perspective, we introduce a method based on reward machines and Hindsight Experience Replay (HER) to translate first-order logic specifications and address reward sparsity. We evaluate this approach to a continuous-control setting using Non-Linear Arithmetic Theory, showing that it enables natural specification of complex tasks. Experimental results show how a tailored implementation of HER is fundamental in solving tasks with complex goals.
\end{abstract}


\section{Introduction}
In Reinforcement Learning (RL), an agent interacts with an environment and receives a scalar reward signal as feedback, which it uses to improve its decision-making over time. In principle, under the \emph{reward hypothesis}~\citep{sutton1998reinforcement}, any task can be modeled by specifying an appropriate reward function. However, reward engineering is a difficult task, often heuristic in nature and reliant on domain knowledge. Furthermore, many tasks of interest are inherently non-Markovian, making commonly used Markovian reward functions less effective for learning. Introducing structure in the reward via logic specification helps address these challenges by bringing reward engineering closer to human language and improving interpretability.
%
In \emph{logic reward specification}, the reward the agent receives at each time step depends on whether a formula is satisfied based on the set of true propositional symbols at that time step. This approach allows expressing complex tasks using \emph{temporal logic}.
To better picture this, consider a robot operating in a warehouse, having to deliver a box to a specific location. This task has two sequential sub-goals: ($A$) picking up the box in a given position $(x_o, y_o)$, then ($B$) delivering it to the designated location $(x_g, y_g)$. This behavior can be enforced via a logic specification that encodes the temporal relationship between events $A$ and $B$, such as "Do $A$, then $B$".

Currently, most works in this line employ logic specification based on Linear Temporal Logic with finite traces (LTLf)~\citep{de2013linear, brafman2018ltlf}. Stemming from its counterpart with infinite traces (LTL \citep{Pnueli77}), which is the most common language for specifying system properties in the field of formal verification. In the last decade, LTLf
has also gained popularity in AI, where reasoning over finite executions, such as in planning tasks, is more intuitive \citep{DeGiacomoR18}. 
LTLf offers sufficient expressiveness for many \emph{discrete} domains, and benefits from a solid toolchain for translating such specifications into automata or ``reward machines"~\citep{icarte2018using}. 
However, the expressiveness of LTLf is limited to Boolean predicates, which constrain the complexity of the reward specification. Evaluating the truth of these predicates typically requires defining \emph{encoding functions} manually. 
%
This limitation is often not an issue in discrete domains or for simple reward functions. However, in many real-world scenarios, such as business process management~\citep{gianola2024linear} or robotics~\citep{li2017reinforcement}, 
a richer reward specification \emph{language} allows providing equally or more informative feedback to the learner \emph{with less human intervention} and can make the specification of new tasks for the same environment significantly faster. 
The importance of an expressive specification language is even more evident in the presence of \emph{heterogeneous} data types, as shown in the following extended example.

\textbf{Warehouse robot example.}
The robot must reach the object's position, identify a specific object by its ID, and satisfy a weight constraint. The resulting predicate might be "$A = (x - x_o)^2 + (y - y_o)^2 < r^2 \wedge i=``\texttt{H123}" \wedge weight < 10$". The robot has a \textit{heterogeneous} state described by variables: $x, y, w \in \mathbb{R}^+$, representing position and object weight ($w = 0$ if no object is being carried) and $i \in \{\texttt{H123}, \texttt{S456}, \ldots\}$ is a given set of IDs. 

Typically, evaluating such a predicate requires some hand-crafted encoders, one to determine when the Euclidean distance between the agent's position $(x, y)$ and the objects location $(x_o, y_o)$ falls below a threshold $r$, and another to check the ID and find the object's weight in an inventory database. Even in this relatively simple example, implementing the specification involves (i) manually encoding each condition, and likely (ii) producing a specification that is not reusable across different tasks. Without such user-defined encodings, handling continuous or heterogeneous states is impossible in LTLf, which operates at the propositional level.
%

To overcome these limitations, we propose using Linear Temporal Logic Modulo Theories over finite traces (LTLfMT) \citep{geatti2022linear,geattiecai,faella2024,Rodriguez024,RodriguezS23,RodriguezAC0K25}, a \ale{well-established} extension of LTLf in which the atomic propositions are replaced by first-order formulas interpreted over (combinations of) arbitrary first-order theories. \ale{This logic was first introduced in \citep{geatti2022linear} to address the limitations of the propositional nature of LTL in modeling complex scenarios, such as those involving arithmetic constraints, rich data types, or relational databases, that go beyond finite-state systems. Such contexts include data-aware processes \citep{HaririCGDM13}, where actions and constraints depend on the content of a persistent data store, typically modelled as a first-order structure like a relational database.} LTLfMT enables reasoning about reward specifications using diverse first-order theories such as arithmetic over the reals and/or integers, uninterpreted functions and relations (to represent, e.g., relational databases \citep{Ghilardi23},\citep{Gianola23}), complex datatypes (e.g., lists, arrays) and even custom domain-specific predicates, providing the expressiveness needed to naturally encode more complex specifications.
From a practical perspective, instead of manually defining separate encoders for each condition, we propose a unified framework that requires only a formula for the task and an assignment of values to constants. The formula can be expressed using different theories and evaluated via a universal Satisfiability Modulo Theories (SMT) solver \citep{BarrettSST21}, without the need to explicitly encode it as a propositional predicate.
Continuing with the delivery robot example, the predicate $A$ discussed earlier can be expressed and evaluated in LTLfMT using the (non-linear) arithmetic theory over the reals, and evaluated via a standard SMT solver without the need for hand-crafted, domain-dependent encodings. By shifting the complexity to the logical specification layer, this framework allows for the definition of more expressive behaviors than LTLf, while maintaining generality and reusability across different tasks and domains though the use of off-the-shelf SMT solvers \citep{CVC5,Z3}.
Theoretically, the LTLfMT framework can be viewed as a composable method for reward specification where each theory acts as a building block, and adding more theories (or combinations thereof) increases the expressiveness of the framework. If no additional theories are included beyond Boolean predicates, we recover the standard LTLf semantics. To achieve greater expressivity, we can incorporate and combine arbitrary theories such as arithmetic and database theories.
%
However, the choice of theories added to LTLfMT (some theories may lack available solvers) and some advanced features of the full logic (such as quantifiers and variable-lookahead operators) affect both decidability and solver support, impacting the framework's efficiency and practical usability. It is thus important to identify sound but sufficiently expressive fragments of the logic and well-supported theories that can handle the data types characteristic of the domain of interest.

\paragraph{Contributions and paper structure.} After discussing related works in Section~\ref{sec:related} and introducing the necessary technical background in Section~\ref{sec:pre}, we present \emph{two main contributions}---
    In \textbf{Section~\ref{sec:logic}}, we define a theoretical framework for reward specifications by extending the classical LTLf logic to the more expressive LTLfMT, in which atomic propositions are replaced with first-order formulas over arbitrary theories. Then, we identify a fragment of LTLfMT that (i) is free from all decidability issues other than those introduced by the theory itself (ii) allows for easy translation of formulas to automata, and yet (iii) is expressive enough to capture most goals of practical interest---
    In \textbf{Section~\ref{sec:rl}} we apply the proposed framework when the integrated theory is
    Non-Linear Real Arithmetic Theory (NRA) to handle reward specifications in a continuous-control scenario. 
    In this domain, we address the problem of reward sparsity that is typical of logic specifications, exploiting the intrinsically ``structured" nature of goals in our first-order framework. 
    We propose a solution that combines Counterfactual Experiences for Reward Machines (CRM) and Hindsight Experience Replay~(HER)~\citep{andrychowicz2017hindsight} in a nontrivial way. 
    Empirical results on continuous control tasks with complex goals suggest an improvement w.r.t. using only CRM (Section~\ref{sec:exp}). The code can be found at: \url{https://bit.ly/4i20C4Z}.


\section{Related Works}\label{sec:related}
We discuss related works in terms of the two key challenges addressed in this paper: dealing with non-boolean data and dealing with reward sparsity.

\paragraph{Propositional languages equipped with labeling functions. }
Given the intrinsic ``discrete" nature of LTL-based rewards, most works focus on MDPs with finite states~\citep{bacchus1996rewarding,thiebaux2002anytime,gretton2003implementation,gretton2011decision,gretton2014more,brafman2018ltlf,bozkurt2020control}. There are notable exceptions:~\citet{cai2021modular} consider continuous MDP states, but logic formulas are defined using propositional letters corresponding to higher level concepts such as ``regions" of the state space. Their definition is extra-logical and left to the user's implementation. This approach is more explicit in the framework of ``restraining bolts"~\citep{de2019foundations,degiacomo2020temporal,degiacomo2020restraining} in terms of ``features" mapping properties of the ``world" (not necessarily observed by the agent) to propositional atoms. When this mapping is applied directly to the states of the MDP, it is usually called \emph{labeling function}~\citep{lacerda2015optimal,hasanbeig2020deep,hasanbeig2023certified,jiang2021temporal}.

Reward Machines (RM)~\citep{icarte2018using,icarte2022reward} are a programmatic alternative to logic specification where the user directly builds a finite-state automaton. In fact, we can identify the automaton generated by an LTLf formula as a type of RM called \emph{simple reward machine} with a binary reward function. 
As in many works on logic discussed above, the mapping between states of the MDP and propositional letters must be implemented separately and is also called a labeling function in this context.

The use of labeling functions has allowed the application of both logically-specified rewards and reward machines to MDPs with continuous states (and actions), opening the door to \emph{deep} RL experiments~\citep[e.g., ][]{hasanbeig2020deep, li2024reward}.
However, in all the aforementioned examples, the labeling function (or equivalent) is treated as a black box. This approach complicates generalization between tasks, is prone to errors, and undermines the interpretability of the logical formulas. 
A notable exception is the temporal logic proposed by \citet{li2017reinforcement} for robotic control, which allows first-order terms of the kind $f(s) < c$, where $s$ is a continuous state of the MDP, $c$ a constant and $f$ a function. We will similarly argue that inequality terms of this kind are enough to encode most goals of interest for the setting of continuous control.
However, the first-order terms of \citet{li2017reinforcement} are mere propositional atoms for which the form of the labeling function is prescribed in advance. 
We will show in which sense LTLfMT is more versatile.

\paragraph{Existing remedies to sparse rewards. }
By definition, goals specified by means of logical formulas tend to generate \emph{sparse} (i.e., rare) rewards. These are notoriously hard to optimize in RL and led to the early development of general-purpose \emph{reward shaping} strategies~\citep{dorigo1994robot,ng1999policy}---the addition of extra feedback to guide the agent towards the goal. Within the recent \emph{deep RL} literature, in particular regarding robotics applications, Hindsight Experience Replay (HER)~\citep{andrychowicz2017hindsight} has established itself as a powerful alternative, if not the standard approach to goal-based tasks. However, HER works by exploiting the state-generalization capabilities of neural networks applied to goals defined as states or functions of state variables. This shared structure of goals and states is typically absent from specifications based on propositional logics or finite-state automata, or hidden inside black-box components such as labeling functions.
The problem of sparsity was explicitly addressed in the RM literature, where several ad-hoc approaches were already proposed by~\citep{icarte2022reward}, namely Counterfactual experiences for Reward Machines (CRM), Hierarchical approaches (HRM), and Automated Reward Shaping (ARS). They also suggested to explore synergies between RM and HER, but to the best of our knowledge the idea was not explored further.
Hierarchical approaches, in the form of temporal abstractions, can be found also in the LTL literature~\citep{araki2021logical}. Notably,~\citet{jothimurugan2019composable} jointly propose a (propositional) reward-specification language and a reward-shaping method.

\section{Preliminaries}\label{sec:pre}
In this section, we review the main concepts of RL and reward specification based on temporal logic. 
\paragraph{Reinforcement learning.} 
The reinforcement learning problem~\citep{sutton1998reinforcement} is usually defined as the solution to a Markov Decision Process (MDP)~\citep{puterman1990markov}. An MDP consists of a tuple $\langle S, A, R, P, \gamma\rangle$ where $S$ is the set of states, $A $ is the set of actions and $R: S \times A \rightarrow \mathbb{R}$ is the reward function. The environment's dynamics are described by the transition function $P : S \times A \times S \rightarrow [0, 1]$, which represents the probability of transitioning to state $s^\prime$ given we are in state $s \in S$ and take action $a \in A$. The discount factor $\gamma \in [0, 1)$ weighs the importance of future rewards versus immediate ones. The agent acts according to a policy $\pi: S \rightarrow \Delta(A)$, which assigns a probability distribution over actions depending on the observed state $s \in S$. 
The optimal policy is the one yielding the largest \emph{return}, that is the expected discounted sum of rewards collected from the initial state following the policy. 
Tasks specifications based on temporal logic or reward machines naturally induce non-Markovian reward functions. This more expressive framework is commonly referred to as a Non-Markovian-Reward Decision Process (NMRDP). Formally an NMRDP is a tuple $\langle S, A, \bar R, P, \gamma \rangle$ equivalent to an MDP except for the non-Markovian reward function $\bar R : (S\times A)^* \rightarrow \mathbb R$, which depends on the entire history of state-action pairs \citep{bacchus1996rewarding}. Classical reinforcement learning methods cannot be applied directly to NMRDPs.

\paragraph{Linear temporal logic over finite traces.}
The LTLf logic \citep{de2013linear} is a variant of LTLf \citep{pnueli1977temporal} defined over finite traces. Logic formulas specified in LTLf are evaluated from propositional Boolean symbols defined in the alphabet $\mathcal{P}$. A finite trace $\sigma = \langle \sigma_0, \sigma_1, \ldots \sigma_n \rangle$ is a sequence of \emph{states} (not to be confused with the states of the MDP). Each state $\sigma_i \subseteq \mathcal P$ is a set of propositional symbols true at the $i$-th timestep, where the finite integer $n \in \mathbb{N}$ specifies the last time step. In LTLf, a formula $\varphi$ is defined by the grammar~
{$\varphi := p \mid \neg \varphi \mid \varphi_1 \wedge \varphi_2 \mid \lnext \varphi \mid \varphi_1 \mathcal U \varphi_2
$
},
where $p \in \mathcal{P}$, $\lnext$ is the \emph{next} operator and $\mathcal U$ is the \textit{until} operator. Other operators can be defined via standard abbreviations, for instance $\varphi_1 \vee \varphi_2 := \neg (\neg \varphi_1 \wedge \neg \varphi_2)$, the \emph{weak next} operator $\lwnext \varphi := \neg \lnext \neg \varphi$ and the \textit{eventually} operator $F\varphi := \top \mathcal{U} \varphi$.
A formula $\varphi$ is evaluated over trace $\sigma$ at a time step $t$, which is formally denoted as $(\sigma, t) \models \varphi$. We adopt the standard semantic of LTLf, defined by induction on the formula~\citep{de2013linear}. The key aspects on top of the semantic of propositional logic are: $(\sigma, t)\models X\varphi$ if and only if $(\sigma, t+1)\models \varphi$, and $(\sigma, t)\models \varphi_1\mathcal{U}\varphi_2$ if and only if $(\sigma, j)\models \varphi_2$ for some $j\ge t$ and $(\sigma, k)\models \varphi_1$ for all $t\le k < j$.
In reward specification based on temporal logic, we interpret propositional symbols in $\mathcal P$ as world features, which the agent receive at each time step $t$. At each time step $t$, the (binary) reward obtained by the agent is decided by evaluating $\varphi$ over the current logical state $\sigma_t$. Typically, the propositional symbols or fluents in $\mathcal{P}$ represent truth values about the MDP state. For example, \cite{de2019foundations} refer to this encoding as a \textit{feature} function $f_i : W \rightarrow D_i$ mapping a world state $w$ from the set of world states $W$ to a finite domain $D$ such as $\{\mathrm{True},\mathrm{False}\}$. A state of the trace $\sigma_t = \langle f_1(w_t), f_2(w_t) \ldots \rangle$ is obtained by extracting the features $f_i$ from the current world state $w_t$ at time $t$. 
The clear non-Markovian nature of this reward signal makes the decision problem a NMRDP.
A widely adopted solution~\citep[e.g.,][]{degiacomo2020restraining} is to convert $\varphi$ into a finite automaton $\mathcal{A}_\varphi = \langle Q, 2^\mathcal{P}, \delta, q_0, F \rangle$, where $Q$ is a set of automaton states, $2^{\mathcal{P}}$ is the input alphabet, $\delta$ is the transition function, $q_0 \in Q$ is the initial state, and $F \subseteq Q$ the set of accepting (terminal) states. The automaton accepts a trace $\sigma = \langle \sigma_0, \sigma_1, \dots, \sigma_n \rangle$ if, starting from $q_0$ and sequentially consuming inputs from $\sigma$, it reaches an accepting state $q \in F$. By combining the NMRDP $\mathcal M$ with the automaton $\mathcal{A}_\varphi$, we construct a product MDP $\mathcal M^\prime := \langle S^\prime, A^\prime, R^\prime, P^\prime, \gamma \rangle$ where the state space $S^\prime = S \times Q$ tracks both the environment state and the automaton state. Transition and rewards in the product MDP are Markovian, as the automaton state capture the necessary information about the history. Hence, $\mathcal{M}^\prime$ is equivalent to the original NMRDP, but can be solved using standard RL techniques.

\section{Reward Specification via LTLfMT}\label{sec:logic}
LTLfMT \citep{geatti2022linear} extends the expressiveness of temporal logic by allowing predicates in $\mathcal{P}$ to be first-order formulas over arbitrary theories. A predicate $p \in \mathcal P$ is no longer a Boolean symbol but can instead represent a first-order formula, interpreted according to a chosen background theory. 
In this section, we formally define the syntax of LTLfMT showing how different theories are integrated into this framework. Then we propose a propositionalizable \emph{fragment} of LTLfMT retaining sufficient expressiveness for reward specification in RL. Finally, we show how to translate a formula of this fragment into a Deterministic Finite Automaton (DFA) that can be used to define a product MDP for reinforcement learning.

\subsection{Linear Temporal Logic Modulo Theories with finite traces}

To represent predicates as first order formulas, in LTLfMT, the alphabet $\mathcal P$ is replaced by a multi-sorted first-order signature $\Sigma = \mathcal{S} \cup \mathcal{P} \cup \mathcal{C} \cup \mathcal{F} \cup \mathcal{V} \cup \mathcal{W}$ where, $\mathcal{S}$ is a set of sort symbols (i.e., data types), $\mathcal C$ is a set of constants, $\mathcal P$ is the set of predicates,\footnote{Note that in LTLfMT there is no alphabet. Hence, from now on  $\mathcal{P}$ denotes the predicate symbols of the first-order signature $\Sigma$.} $\mathcal F$ the set of functions, $\mathcal V$ and $\mathcal W$ are the sets of variables and quantified variables respectively. A $\Sigma$-term $t$ serves as a building block for constructing atomic formulas in LTLfMT, and is generated by the following grammar:
\begin{equation}
    \label{eq:grammar_sigma_term}
	t := v \mid w \mid c \mid f(t_1, \ldots, t_k) \mid \lnextmt v \mid \lwnextmt v 
\end{equation}
where $c \in \mathcal{C}$, $v \in \mathcal{V}$, $w \in \mathcal{W}$ and $f \in \mathcal{F}$ is a function symbol of arity $k$. The operators $\lnextmt$ and $\lwnextmt$ denote the lookahead and weak lookahead operators, respectively, similar to the next $\lnext$ and weak next $\lwnext$ in LTLf, but acting on $\Sigma$-variables; they are used to relate the value of variables at a given time to their value at the following instant. The complete grammar of LTLfMT is defined as:
\begin{equation}
\label{eq:grammar_ltlfmt}
\begin{aligned}
        \alpha &:= p(t_1, \ldots, t_k) ,\\
    \lambda &:= \alpha \mid \neg \alpha \mid \lambda_1 \vee \lambda_2 \mid \lambda_1 \wedge \lambda_2 \mid \exists v \lambda \mid \forall v \lambda,\\
    \varphi &:= \top \mid \lambda \mid \varphi_1 \vee \varphi_2 \mid \varphi_1 \wedge \varphi_2 \mid \lnext \varphi \mid \lwnext \varphi \mid \varphi_1 \mathcal{U} \varphi_2. 
    \end{aligned}
\end{equation}
Compared to LTLf, this grammar introduces two additional layers $\alpha$ and $\lambda$. Intuitively, $\alpha$ defines atomic predicates over terms, and $\lambda$ allows first-order combinations over these atoms. The top-layer formulas $\varphi$ are standard temporal formulas over these first-order expressions, similar to LTLf.
%
Theories are integrated into this framework via the interpretation of $\Sigma$-terms. A theory $\mathcal{T}$ can be defined as a set of $\Sigma$-structures, i.e., sets that interpret all the symbols in $\Sigma$ (so, also $\Sigma$-terms) which are called models of the theory. Intuitively, a $\Sigma$-signature defines the available symbols (e.g., $1, 2, +, >, \ldots$), a $\Sigma$-structure provides a concrete interpretation (e.g., $+(1, 2)$ is interpreted as integer addition
). A theory $\mathcal T$ provides interpretations for each symbol in $\Sigma$. For example, the Linear Integer Arithmetic (LIA) theory introduces a \emph{sort} for \texttt{Integer} data types into $\mathcal{S}$, arithmetic operations such as $+$ and its inverse $-$ in the set $\mathcal{F}$, constants and variables valued in $\mathbb Z$ via $\mathcal{C}$, $\mathcal{V}$ and $\mathcal{W}$, and comparison predicates (e.g., $=$, $<$) into $\mathcal P$. 
Multiple theories can be combined, including their $\Sigma$-signatures, and their interpretations are given through combined $\Sigma$-structures.
Formally a $\mathcal T$-state $s = (M, \mu)$, contains two elements, $M \in \mathcal{T}$ (i.e., a $\mathcal T$-model) is a first-order $\Sigma$-structure that interprets all the symbols in $\Sigma$ and $\mu: \mathcal{V} \rightarrow \text{dom}(M)$ a function mapping non-quantified variables to their domain of reference. Quantified variables are interpreted by a separate environment function $\xi : \mathcal W \rightarrow \text{dom}(M)$. A trace (or word), in LTLfMT is a finite sequence of $\mathcal T$-state over time-steps, $\sigma = \langle (M, \mu_0), \ldots, (M, \mu_n)\rangle$ where $M \in \mathcal T$ is fixed across the trace, and $\mu_i$ may differ, reflecting the evolution of variables over time. The evaluation of a term $t$ at time $i$ in a trace $\sigma$ under variable environment $\xi$ is defined using the functions $\mu_i$ and $\xi$ in the base case of variables in $\mathcal V$ and $\mathcal W$, resp., and defined inductively in the usual way for constants and functional terms.
The satisfaction relation for an LTLfMT formula $\varphi$ over a trace $\sigma$, at time-step $i$ with environment function $\xi$ is denoted as $\sigma, \xi, i \models \varphi$ and follows the usual definition of first-order semantics. This satisfaction relation is used to define the \emph{satisfaction modulo $\mathcal T$} of an LTLfMT formula $\varphi$ over the trace $\sigma$ at time point time-step $i$, denoted as $\sigma, i \models \varphi$ and inductively defined in the standard way to deal with the temporal operators (see \citep{geatti2022linear} for the formal definition of the semantics).


\subsection{Fragment for Reward Specification in RL}\label{sec:frag}
LTLfMT framework allows for very general expressions, and it is in general undecidable \citep{geatti2022linear}. It is possible to define a fragment of LTLfMT that is semi-decidable. Formally, for
decidable underlying first-order theories, LTLfMT has been proven to be semi-decidable (i.e., a positive answer can always be obtained for satisfiable
instances, while reasoning might not terminate over unsatisfiable instances), thanks to the semi-decision procedure shown by \citet{geatti2022linear}, and also implemented in the BLACK solver. \footnote{https://www.black-sat.org/en/stable/}
Intuitively, for identifying a decidable fragment of LTLfMT, the underlying theory $\mathcal T$ must necessarily be decidable, but this is not sufficient in general: decidability in first-order extensions of LTLf may still be broken by the interaction of temporal operators and quantifiers, i.e., in the case of LTLfMT, by the presence of the lookaheads operators.
We propose a tractable fragment of LTLfMT for reward specification in reinforcement learning by removing the (weak) lookahead operator $\lnextmt$ ($\lwnextmt$) from the $\Sigma$-layer. While the (weak) next operator $\lnext$ ($\lwnext$) in the temporal logic layer is essential for specifying temporal constraints between ``subtasks'' that the agent may have to complete, the addition of (weak) lookahead operator on the $\Sigma$-term, $\lnextmt$ ($\lwnextmt$), which applies to variables, is neither natural nor beneficial for reward specification. The $\Sigma$-layer can be regarded as a data collection interface for capturing unstructured and heterogenous environmental information. We argue that defining temporal constraints on data collection is not useful for reward specification, and that the $\Sigma$-layer should just be used to capture immediate information available to the agent. 
Note that our fragment of LTLfMT still supports arbitrary first-order theories. 
\begin{align}\label{eq:grammar_lfltlfmt}
	t &:= v \mid w \mid c \mid f(t_1, \ldots, t_k)\nonumber\\
    \alpha &:= p(t_1, \ldots, t_k) \nonumber\\
    \lambda &:= \alpha \mid \neg \alpha \mid \lambda_1 \vee \lambda_2 \mid \lambda_1 \wedge \lambda_2 \mid \exists v \lambda \mid \forall v \lambda\\
    \varphi &:= \top \mid \lambda \mid \varphi_1 \vee \varphi_2 \mid \varphi_1 \wedge \varphi_2 \mid \lnext \varphi \mid \lwnext \varphi \mid \varphi_1 \mathcal U \varphi_2 \nonumber
\end{align}

In this fragment, the source of undecidability becomes the theory itself (and the use of quantifiers). Assuming the theory itself is decidable, we can leverage any tool implementing a decision procedure for the theory itself to deal with the first-order dimension: this is the case if the quantifier-free fragment of the theory is decidable and the quantifiers do not affect decidability because, for instance, a quantifier elimination procedure is available (as in the case of real arithmetic \citep{Tarski}). In such scenarios, we can employ a state-of-the-art QE module for eliminating the quantifiers \citep{Collins74}, and then universal tools such as SMT solvers \citep{BarrettSST21} to evaluate the obtained quantifier-free formula. Depending on the theory, in cases where the use of quantifiers leads to undecidability, we can further restrict our fragment by removing them. Doing so, we obtain a LTLfMT fragment similar to ``data-LTLf'' by \citet{gianola2024linear}.

\subsection{Translation to DFA and Product MDP} 
\citet{geatti2022linear} give a non-constructive proof that a Boolean abstraction of any language definable in lookahead-free LTLfMT can also be defined in LTLf, and vice versa.
We take a more practical viewpoint and provide a propositionalization of our fragment in the following sense: (i) a syntactic transformation of lookaead-free LTLfMT formulas into LTLf formulas, and (ii) a transformation of LTLfMT traces into LTLf traces. This is enough to define an automaton and a product MDP that can be simulated efficiently using only a theory solver for the elected theory (such as an SMT solver if the theory is quantifier-free).
Refer to the grammar of our fragment in Equation~\eqref{eq:grammar_lfltlfmt}.
Given a formula $\varphi$ from our language, we first substitute each $\alpha$ containing only free variables from $\mathcal{V}$ with a distinct propositional letter. Then, we substitute each remaining $\lambda$ with a distinct propositional letter. The result is a LTLf formula $\varphi'$.
Let $\mathcal{P}$ now denote the set of all the employed propositional letters---our new alphabet.
A LTLfMT trace $\sigma=(s_0,\dots,s_t)$ is mapped to an LTLf trace $\sigma'=(\sigma'_1,\dots,\sigma'_n)$ by assigning a truth value to each propositional letter in $\mathcal{P}$ at each step. This is done, e.g., with a theorem prover \citep{Vampire} or a decision procedure for the theory of choice (or an SMT solver if the theory is quantifier-free, as detailed in Section~\ref{sec:frag}).
Crucially, the problem given to the solver at a given timestep is purely first-order, without any temporal operator.
For example, assuming the theory is real arithmetic, if "{$\forall y \exists z (y*z>0 \implies z<y)$}" from $\phi$ was replaced with letter $P$ in $\phi'$, the solver decides $P\in \sigma'_t$ if and only if $s_t\models \forall y \exists z (y*z>0 \implies z<y)$, applying efficient quantifier elimination first.
At runtime, the propositional abstraction $\varphi'$ is evaluated on the $LTLf$ trace $\sigma'$. The SMT solver plays the role of an extremely general \emph{labeling function} (cf. Section~\ref{sec:related}).
For an RL-friendly implementation, we can translate $\varphi'$ into a Deterministic Finite Automaton (DFA) $\mathcal{A}_{\varphi'} = \langle Q, 2^\mathcal{P}, \delta, q_0, F \rangle$ and use it to define a product MDP $M^\prime := \langle S^\prime, A^\prime, R^\prime, P^\prime, \gamma \rangle$ with extended state space $S^\prime = S \times Q$, exactly like in LTLf (cf. Section~\ref{sec:pre}).
We can do this automatically using the translation technique defined in \cite{zhu2019firstordervssecondorderencodings}\footnote{In our implementation, we use the \texttt{LTLf2DFA} library (\url{https://github.com/whitemech/LTLf2DFA})}
At training time, state transitions in the DFA are resolved by the SMT solver. This is a special kind of reward machine (specifically, a \emph{simple} reward machine with a Boolean-valued reward function, cf.~\citep{icarte2018using}), again with the SMT solver playing the role of the labeling function.

\section{Reinforcement Learning with LTLfMT Rewards}\label{sec:rl}

In this section, we provide an RL solution to the tasks defined using our proposed specification language, with a focus on the continuous control-domain.

Let us first recap our reward specification pipeline, stressing that it reduces significantly manual interventions compared to other methods.  
The end goal of the specification is to construct a product MDP to which standard RL algorithms (and more sophisticated techniques inspired by reward machines) can be applied. First, depending on the domain of interest (e.g., continuous control), we pick a theory (e.g., NRA) with an associated SMT solver (with all the precautions of Section~\ref{sec:frag}). Several different tasks can be specified within the domain of interest.
From the user defining the task, or \emph{reward engineer}, the only input we require is a formula $\varphi$ written in the lookahead-free LTLfMT fragment defined in Section~\ref{sec:frag}. 
Compared to existing reward specification frameworks (cf. Section~\ref{sec:related}), the user need not manually implement a labeling function or anything similar.
At most, to enable \emph{multi-goal RL} (for example, to use HER as we will detail later) the user needs to provide a dictionary assignment of first-order constants appearing in the formula to numerical values. In this way, the task is "parametrized" by (a subset of) the constants. In the following, we will focus on continuous control, both to provide an intuitive, realistic example of our reward specification pipeline and study the effectiveness of HER in dealing with sparse rewards generated by logic specifications.


\paragraph{Continuous Control Case Study}

Common tasks in continuous control consist in controlling a variable to a target value or range, or a vector of variables to a target point. For example, a target position in space for a robot, a desired temperature for a HVAC system, a speed for cruise control.
These "subtasks" can be described in terms of $L^p$ distances between state vectors (e.g., Euclidean distance, box contraints) This can be done using real arithmetic, and subtasks can be combined into more complex tasks using temporal operators. 
Hence, for continuous control, we can select Non-Linear Real Arithmetic (NRA) as our theory for lookahead-free LTLfMT, with all the computational advantages discussed in Section~\ref{sec:frag}.
For example, consider an autonomous car moving in $2$-dimensional space within a parking area. A state $s \in \mathbb{R}^2$ represents spatial coordinates. We specify a (parametric) task where the agent must reach a point $A$ in space first, then a point $B$, without ever going through a certain "unsafe" area. The task can be specified in our lookahead-free LTLfMT fragment as $\varphi = \neg F \neg (x \geq x_{min} \wedge x \leq x_{max}) \wedge F((x - x_a)^2 + (y - y_a)^2 < r_a^2 \wedge F((x - x_b)^2 + (y - y_b)^2 < r_b^2))$. 
In this case, we use an SMT solver suited for Linear Real Arithmetic (LRA) and Non-Linear Real Arithmetic (NRA) to propositionalize $\varphi$ and obtain $\varphi^\prime := \neg F \neg \alpha_1 \wedge F(\alpha_2 \wedge F(\alpha_3)$.

We use this example to stress the advantages of our reward specification framework. First, the input we require to the user to specify the task is very simple. It consists of a simple string representing the specification formula $\varphi$, along with an optional assignment of constants to numerical values to parametrize the task.\footnote{This can be as simple as a \texttt{json} configuration file specifying the names of the constants and the values.} 
In contrast, existing approaches often require complex inputs. For example, reward machines as proposed by~\citep{icarte2018using} require explicit specification of the automaton. They also require an implementation of the labeling function, just as several logic-based approaches discussed in~\ref{sec:related}.
Moreover, defining new tasks for the same domain can be very quick in our framework. In the continuous control domain we have outlined, NRA allows defining many new tasks by \emph{just} writing new formulas. Even more substantial modifications to the domain just require changing the theory (for example, from NRA to linear arithmetic), which could be almost effortless depending on the availability of efficient solvers. 
Finally, in the case of continuous control, our framework makes it easy to handle sparse rewards with HER, as shown in the next section.
\mat{questo paragrafo è utile ma un po' ridondante, e ci serve spazio...}

 \subsection{Addressing Sparsity in the Product MDP}\label{sec:sparsity}
A well known issue of logic-based reward specification methods is the sparsity of rewards. 
For the continuous-control domain, we propose to combine the idea of counterfactual reasoning (CRM) from reward machines~\citep{icarte2022reward} with Hindsight Experience Replay (HER). 
Both techniques work at the data collection level, generating artificial experiences to improve learning. Both can be applied to trajectories generated from the product MDP $\tau = \{<s_0, q_0>, a_0, r_1, <s_1, q_1>, \ldots, <s_T, q_T>\}$. Counterfactual reasoning exploits the underlying reward automaton $\mathcal{A}_{\varphi}$, producing additional "fake" experience by replacing the DFA state with other possible ones. This is always possible since the DFA, differently from the MDP, is simulated by the agent itself. Namely, given a single experience $\{\langle s_t, q_t \rangle, a_t, r_{t+1}, \langle s_{t+1}, q_{t+1} \rangle\}$, we can create $|Q|$ artificial transitions replacing $q_t$ with any other $q\in Q$ and recomputing the reward. 
HER is designed for goal-based scenarios with sparse binary reward feedback. It improves sample efficiency by learning from unsuccessful trajectories, exploiting the state-generalization capabilities of neural networks. HER requires the definition of a "goal-based problem" consisting of a goal space $\mathcal G$, a predicate reward function $f_g: S\rightarrow \{0, 1\}$ that checks whether goal $g \in \mathcal G$ is achieved in state $s \in S$, and a mapping $m : S \rightarrow \mathcal{G}$ s.t. $\forall s \in S, f_{m(s)}(s) = 1$ used to generate artificial goals $g^\prime = m(s)$. A goal-based transition is a tuple $((s_t, g), a_t, r_{t+1}, (s_{t+1}, g))$. 
First, we show how to adapt CRM and HER to our framework, then we combine them to exploit their synergy
 to mitigate reward sparsity.   


\paragraph{Adapting HER to the product MDP.}
HER requires a goal specification, which is not intuitive to define over the states of the product MDP and is usually provided by the user. To automate HER, we define as "true" goal of the product MDP the state $\langle s, q \rangle \in S^\prime$, where $q$ is the predecessor of an accepting state $q_f \in F$ on the automaton $\mathcal{A}_{\varphi^\prime}$ and $s$ is such that $q_f = \delta(q, L(s, \cdot, \cdot))$, that is a state triggering the transition to $q_f$. 
\mat{commentare sul perchè definiamo il goal in questo modo e se questa definizione è unica o no}
Then, the goal-based specification can be defined as follows: $\mathcal{G} = S^\prime$, $f_g(s) = \mathbb{I}\left[m(s) = g\right]$, and $m(s) = s$, where $s \in S^\prime$ is a state of the product MDP.
\mat{qui nel mapping stato-goal non dovremmo sfruttare l'assegnamento delle costanti, il famoso json? Importante sottolineare qua che può essere fatto tutto in modo automatico}\pier{In teoria le costanti del json servono sono nella fase pre SMT dove sono sostituite alla formula parametrica, potremmo pensarci se possono essere utili in questo punto}

\begin{figure*}[t]
    \includegraphics[width=\linewidth]{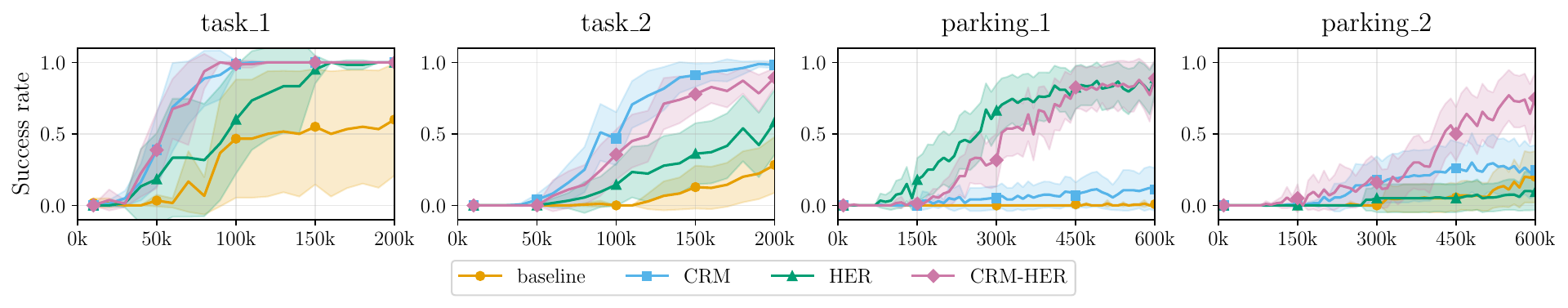}
    \caption{Success rate of each baselines for parking environment in 4 different tasks, with 95\% bootstrap confidence interval over 20 independent experiments. The x-axis reports the training steps of DDPG. Task names are specified above each plot and are ordered left-to-right by task complexity.}
    \label{fig:parking_results}
\end{figure*}

\paragraph{Combining CRM and HER.}
Given a trajectory $\tau$, we add to the replay buffer \mat{abbiamo mai menzionato il replay buffer? Strano anche che non abbiamo mai parlato di DDPG o comunque di un generico algoritmo RL per il continuo fin qui} $|Q|$ artificial experiences obtained by copying $\tau$ and applying counterfactual reasoning as described above. Then, we apply HER to each (real or counterfactual) trajectory generating a total of $2|Q|$ artificial experiences,
\mat{spiegare perchè 2|Q|}
all added to the replay buffer. The intermediate goals of HER are defined using the last state visited in each trajectory $\tau$ \citep{andrychowicz2017hindsight}. \mat{fare riferimento al paper HER per questa scelta}
In the next section we compare both the separate and combined approaches.

\section{Experiments}\label{sec:exp}
In this section we use Deep Deterministic Policy Gradients (DDPG)~\citep{lillicrap2016continuous} to learn a policy from the product MDP constructed for the \textit{parking environment}. We use as baselines different approaches, operating at the \emph{replay buffer} level, presented in Section~\ref{sec:sparsity}.  
%
    \textbf{Baseline}: the task is learned directly from the product MDP using plain DDPG. 
    \textbf{CRM}: CRM is applied to the DFA that specifies the task generating virtual experiences for DDPG.
    \textbf{HER}: HER is applied directly to the augmented state of the product MDP, without using CRM.
    \textbf{CRM-HER}: our main proposed approach, where HER is applied to real experiences and counterfactual experiences generated with CRM.
Our goal is to assess how HER and CRM-HER compare to prior methods across tasks of varying complexity, and whether the supposed synergy of CRM and HER translates into practical advantages.\mat{ho aggiunto questo secondo obiettivo}

\paragraph{Environments and tasks.}
We conducted experiments on a range of tasks (i.e., different $\varphi$ formulae) defined within two distinct continuous control environments: \textit{Parking} and \textit{Reacher}.\mat{controlliamo di aver messo in chiaro in qualche punto del paper la differenza tra "environment" e "task"}
The \textit{Parking} environment, from the \textit{HighwayEnv} collection \citep{highway-env}, involves an agent controlling a car in an empty parking lot. We modified this environment to accommodate task specification defined by arbitrary LTLfMT-NRA formulas. 
%
In each of the two underlying environments, we defined multiple tasks of increasing complexity using our specification pipeline detailed in Section~\ref{sec:rl}.
The complexity of these tasks stems from two main factors: the size and structure of the formula used to define the task (reflected in the complexity of the resulting DFA/product MDP) and the difficulty of discovering the goal, which becomes particularly challenging due to the presence of sparse rewards. Complete details about task specification (including LTLfMT formulas) and results for \emph{Reacher} are in appendix.


\paragraph{Analysis of results.}
Figure~\ref{fig:parking_results} presents the results obtained by averaging performance over multiple training runs for each method. Performance is measured as the \emph{success rate} during evaluation episodes conducted periodically throughout training. An episode is considered successful if the agent reaches the goal within the episode's time limit.
Our results show that, in most tasks, the baseline method struggles to consistently learn a policy capable of reaching the objective, lagging behind the other approaches.
This trend holds across all difficulty levels but becomes especially pronounced in the more complex tasks, where it completely fails to discover a successful policy (e.g., see results for tasks \texttt{parking\_1}, \texttt{parking\_2}).
In most cases, CRM is more effective at handling the increase in complexity of the DFA. This is particularly evident in \texttt{tasks\_1} and \texttt{task\_2} of the Parking environment \fausto{refereces}, where its performance is on par with the best-performing methods.
However, when the goal is difficult to discover, CRM alone offers little benefit, and its performance falls behind that of HER-based approaches (see especially \texttt{parking\_1}).
%
%
RM-HER exhibits mixed performance, with results varying depending on the nature of the task. In tasks where the primary challenge lies in goal discovery, it performs well, clearly demonstrating the benefits of HER (e.g., see \texttt{task\_2}, \texttt{parking\_2}).
In other tasks, however, the improvement over the baseline is less pronounced or, in some cases, barely noticeable.
CRM-HER consistently performs well across all tasks, either achieving the best results or coming close to the top-performing method.
The results indicate that CRM-HER effectively combines the strengths of both CRM and HER, enabling it to handle increasing task complexity. The performance advantage becomes especially clear in the more challenging tasks—most notably in \texttt{parking\_2} task, where it is the only method able to learn a successful policy. The code and the experimental results can be found at: \url{https://bit.ly/4i20C4Z}.

\section{Conclusion}
We proposed a new framework for reward specification in RL based on a fragment of the LTLfMT logic. Compared to LTLf, our language is more expressive as it allows predicates to be first-order logic formulas over arbitrary theories. Although the definition of the logic-augmented MDP follows the standard procedure of LTLf, the increased expressiveness of the language has two important consequences for the user responsible of specifying the reward: (i) the user is relieved of the burden of \emph{coding} labeling functions, which are replaced by off-the shelf SMT solvers, only having to \emph{write a logical formula} and (ii) in continuous-control domains, with minimal effort, they can provide a goal specification for HER using the same variables used in the formula. 
We exploit this second advantage to address the inevitable sparsity of the rewards resulting from such a logic-based specification, combining the CRM technique of reward machines with HER. Experiments in benchmark continous-control domains on tasks specified with lookahead-free LTLf-modulo-NRA show promising results. Future work should explore integration of these techniques with hierarchical RL methods~\citep{icarte2022reward}.
Finally, our reincorporation of first-order terms \emph{inside} the specification language paves the way to future further combinations of symbolic aspects with \emph{continous-space} RL, such as the specification of safety constraints, of continuous rewards, of multi-task problems, ultimately to the formal verification of learning processes in continuous domains.

\section*{Acknowledgments}
This work was partially supported by the `OptiGov' project, with ref. n. 2024.07385.IACDC (DOI: 10.54499/2024.07385.IACDC), fully funded by the `Plano de Recuperação e Resiliência' (PRR) under the investment `RE-C05-i08 - Ciência Mais Digital' (measure `RE-C05-i08.m04'), framed within the financing agreement signed between the `Estrutura de Missão Recuperar Portugal' (EMRP) and Fundação para a Ciência e a Tecnologia, I.P. (FCT) as an intermediary beneficiary. This work was also partly supported by Portuguese national funds through Fundação para a Ciência e a Tecnologia, I.P. (FCT), under projects UID/50021/2025 and UID/PRR/50021/2025.

\bibliographystyle{plainnat}
\bibliography{references}


\appendix

\section{Task Descriptions}
In this section, we provide additional details about the experimental setup. We describe the environments we used and the tasks (both in formula and text description) we defined for our experiments.
Assignments of numerical values to certain constants display our manual configuration of the tasks (goal assignment), but also which constants HER can manipulate to define fictitious goals.

\subsection{Parking Environment}
The parking environment \citep{highway-env} represents a two-dimensional parking lot where the agent controls a car, represented by a continuous vector $s \in \mathbb{R}^6$. The state contains information about spatial coordinates $x, y$, velocities $v_x, v_y$, and orientation $\alpha = \sin\theta, \beta = \cos{\theta}$ where $\theta$ is the angle of yaw. 
The action space is continuous and $2$-dimensional (throttle and steering angle).
In the following, we list the tasks we defined for this environment. The only variables appearing in the formulas are $x,y,\alpha,\beta$, while $a,b,c,d$ are constants.

\paragraph{Task 1}
In \texttt{task\_1} the agent is required to first navigate a checkpoint position $A=(-0.2, -0.08)$, then to move toward a goal position $G=(a, b)$ in the parking lot, respecting this temporal order. The formula describing this task consists of a distance constraint on the spatial coordinate variables of the car $(x, y)$, with a tolerance of $r=0.03$:

\begin{equation*}
    \varphi_1 = \ F ( \ (x + 0.2)^2 + (y +0.08)^2 < 0.03^2 \ \land \lnext F \ (x - a)^2 + (y - b)^2 < 0.03^2 \ ).
\end{equation*}

Goal assignment: $\{a=0.2, b=0.08\}$. 

\paragraph{Task 2} 
In \texttt{task\_2}, we introduce an additional target position $B=(e, f)$. The agent is required to navigate both checkpoints $A, B$ in any order and then move to the goal $G$. Below the corresponding formula:
\begin{align*}
    \varphi_2 = \ & F ( \ (x +0.2)^2 + (y +0.08)^2 < 0.03^2 \ \land \lnext \\ 
    & F( \ (x - 0.2)^2 + (y +0.08)^2 < 0.03^2 \ \land \lnext F \ (x - a)^2 + (y - b)^2 < 0.03^2 \ ) \ ) \\
    & \lor \\
    & F ( \ (x - 0.2)^2 + (y +0.08)^2 < 0.03^2 \ \land \lnext \\
    & F( \ (x +0.2)^2 + (y +0.08)^2 < 0.03^2 \ \land \lnext F \ (x - a)^2 + (y - b)^2 < 0.03^2 \ ) \ ).
\end{align*}
Goal assignment: $\{a=0.2, b=0.08\}$. 

\paragraph{Task 3} 
This task, denoted in Figure~\ref{fig:parking_results} as \texttt{parking\_1} introduces a stricter specification for the goal $G$.
After going through checkpoint A, instead of just reaching a position in the parking lot, the agent is required to correctly park the car in a specific spot, which means that the car must also meet orientation constraints. This is formalized via a box constraint (weighted $L_1$ norm) over spatial and orientation variables $x, y, \alpha, \beta$ as follows:

\begin{align*}
    \varphi_3 =  & F \ ( \ (x +0.2)^2 + (y +0.08)^2 < 0.03^2 \ \land \lnext  \\
    & F \ (|x - a| + 0.2|y - b| + 0.02|\alpha-c| + 0.02|\beta - d| < 0.0144
    \ ) \ ).
\end{align*}
where $r=0.0144$ is the tolerance threshold.
Goal assignment: $\{a=0.18, b=0.14,c=0, d=1\}$. 

\paragraph{Task 4}
This task referred to as \texttt{parking\_2} is the most difficult in our tests, as it combines the complexity of previous tasks. It requires the agent to pass through the two checkpoint positions $A, B$ (in any order), then to correctly park the car in a specific spot.
\begin{align*}
    \varphi_4 = \ & F ( \ (x +0.2)^2 + (y +0.08)^2 < 0.03^2 \ \land \lnext F( \ (x -0.2)^2 + (y +0.08)^2 < 0.03^2 \ \land \lnext \\
    & F \ (
    |x - a| + 0.2|y - b| + 0.02|\alpha-c| + 0.02|\beta - d| < 0.0144
    \ ) \ ) \\
    & \lor \\
    & F ( \ (x - 0.2)^2 + (y +0.08)^2 < 0.03^2 \ \land \lnext F( \ (x +0.2)^2 + (y +0.08)^2 < 0.03^2 \ \land \lnext \\
    & F \ (
    |x - a| + 0.2|y - b| + 0.02|\alpha-c| + 0.02|\beta - d| < 0.0144
    \ ) \ ).
\end{align*}
Goal assignment: $\{a=0.18, b=0.14, c=0, d=1\}$. 

\paragraph{Analysis of results for parking}
Additional results for parking are shown in Figure~\ref{fig:reacher_all}, where for each task we report cumulative regret and cumulative reward over 20 runs. The cumulative regret is measured against an ideal policy that always succeeds. These additional results reflect the analysis made in Section~\ref{sec:exp}.

\begin{figure}
    \centering
    \includegraphics[width=\linewidth]{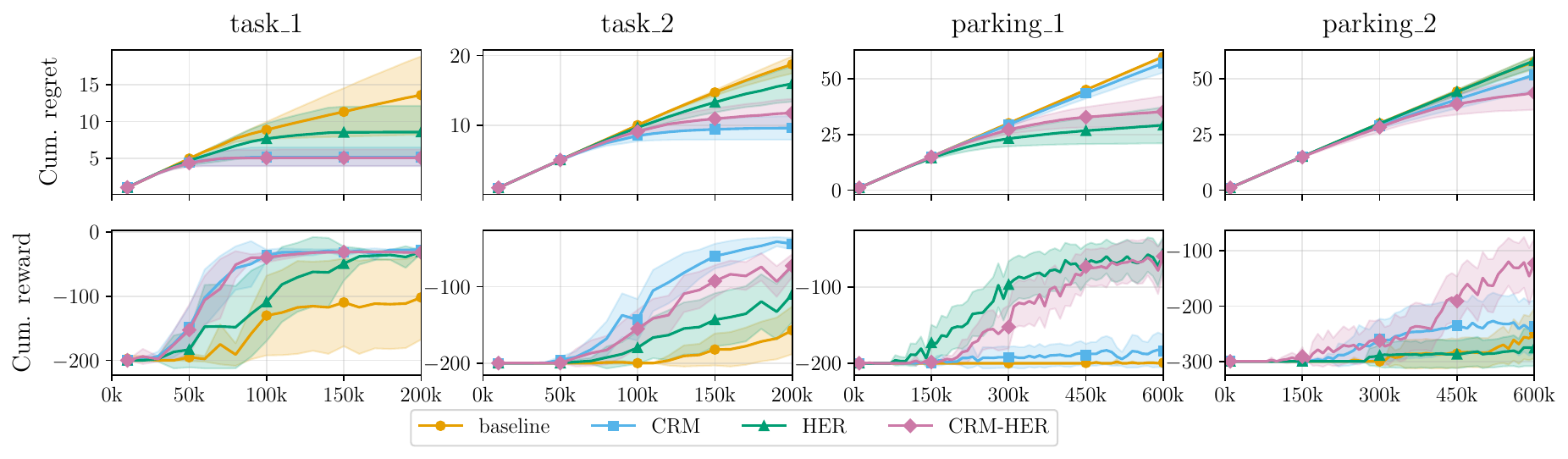}
    \caption{Regret and cumulative reward plot for each task on Parking environment.}
    \label{fig:parking_reg_rew}
\end{figure}

\subsection{Reacher Environment}
Reacher is a two-jointed robotic arm moving in a two-dimensional space. 
We use a variant of \texttt{Reacher-v5}\footnote{\url{https://gymnasium.farama.org/environments/mujoco/reacher/}} with a modified state space. 
Our state $s \in \mathbb{R}^{11}$ encodes the sine and cosine of the angle of each arm $\alpha_1,\beta_1, \alpha_2, \beta_2$, the angular velocity of each arm $\omega_1, \omega_2$, the spatial position of the target $(x_t, y_t)$, and the coordinates of the arm hand $(x, y, z=0)$.
The action space is continuous and $2$-dimensional (torques of two hinges).
Below we describe the tasks we defined for reacher. The only variables appearing in the formulas are $x$ and $y$, while $a$ and $b$ are constants.

\begin{figure}
    \centering
    \includegraphics[width=\linewidth]{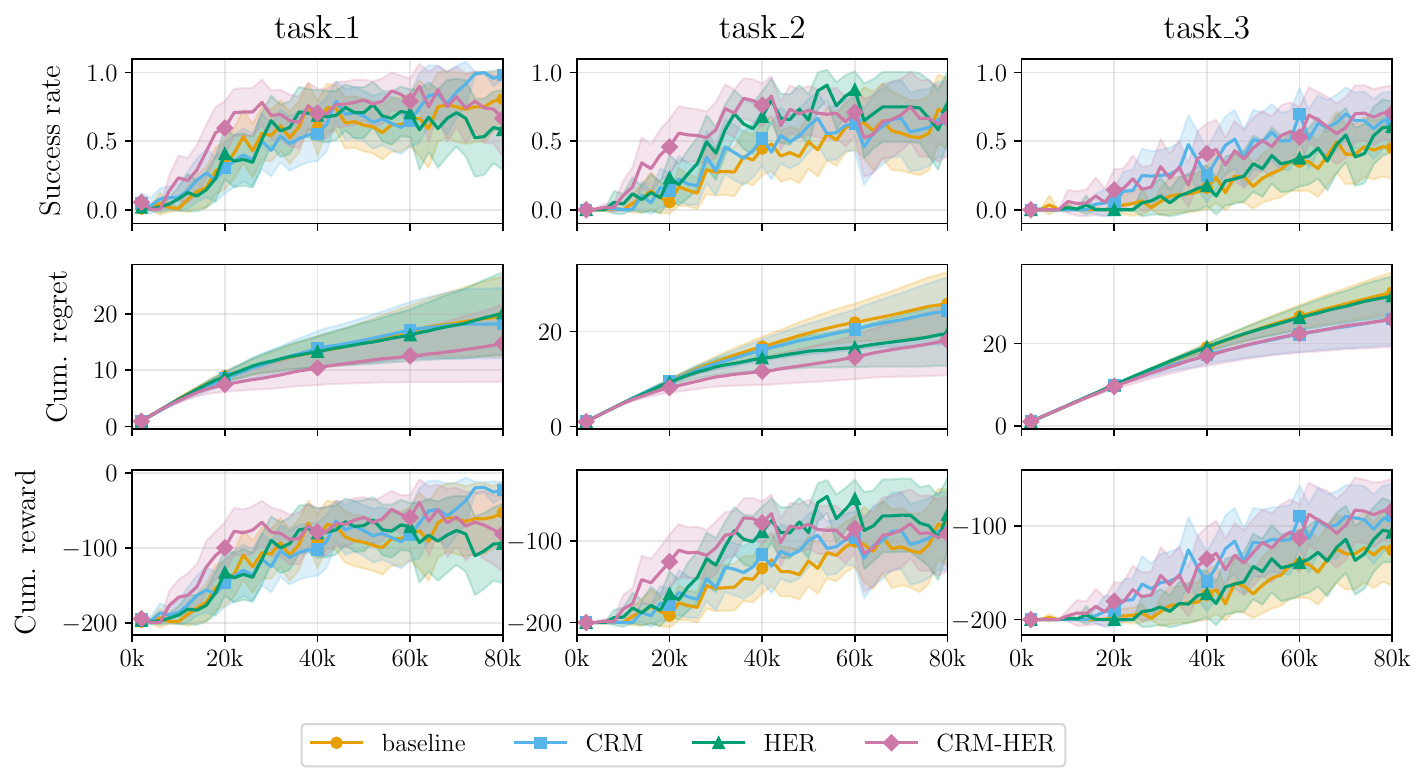}
    \caption{Success rate, cumulative regret and cumulative reward on each task for the reacher environment.}
    \label{fig:reacher_all}
\end{figure}

\paragraph{Task 1}
In the task named \texttt{task\_1} in Figure~\ref{fig:reacher_all}, the agent must move the hand $(x, y)$ to the left, which means moving in any position described by the box constraint $x < -0.19$, then move it to a specified goal position $G=(a,b)$, within $r_g=0.01$ tolerance radius. The formula is as follows:

\begin{equation*}
    \varphi_1 = F ( \ x \leq -0.19 \ \land \lnext F \ (x - a)^2 + (y -b)^2 < 0.01^2 \ ).
\end{equation*}
Goal assignment: $\{a=0.1, b=0.1\}$.

\paragraph{Task 2}
Task \texttt{task\_2} requires an additional step. After moving the arm to the left, the agent has to move the hand close to it's base $S=(0, 0)$, with a tolerance distance of $r_s=0.03$, only then it must reach the goal position $G$. The formula is as follows:
\begin{equation*}
    \varphi_2 = F ( \ x \leq -0.19 \ \land \lnext F \ ( \ x^2 + y^2 < 0.03^2 \ \land \lnext F \ (x - a)^2 + (y - b)^2 < 0.01^2 \ ) ).
\end{equation*}
Goal assignment: $\{a=0.1, b=0.1\}$.

\paragraph{Task 3}
Task \texttt{task\_3} is similar to \texttt{task\_2}, with a stricter tolerance $r_g=0.     005$ to reach the goal position, reducing the tolerance distance from $0.01$ to $0.005$, which increases the complexity significantly reducing the accepted goal area. The formula, is almost equal to previous task:
\begin{equation*}
    \varphi_3 = F ( \ x \leq -0.19 \ \land \lnext F \ ( \ x^2 + y^2 < 0.03^2 \ \land \lnext F \ (x - a)^2 + (y - b)^2 < 0.005^2 \ ) )
\end{equation*}
Goal assignment: $\{a=0.1, b=0.1\}$.

\paragraph{Analysis of results for reacher}
Results for the Reacher environment are shown in Figure~\ref{fig:reacher_all}. As before, we compare success rate, cumulative regret and cumulative reward across 20 runs. While no single technique strictly dominates in terms of success rate or cumulative reward, CRM-HER consistently performs comparably or better than other approaches and demonstrates faster convergence in tasks \texttt{task\_1} and \texttt{task\_2}. This improvement becomes even more evident when analyzing cumulative regret, where CRM-HER emerges as a consistently preferable choice over other methods.

\section{Hyperparameters}

\begin{table}
    \textbf{Parking Environment Hyperparameters}
    \centering 
    \small
    
    \begin{tabular}{|c | c | c | c | c|}
    \hline
     & \textbf{Task 1} & \textbf{Task 2} & \textbf{Task 3} & \textbf{Taks 4} \\
    \hline \hline
    \textbf{episode length} & 200 & 200 & 200 & 300\\
    \textbf{buffer size} & $10^6$ & $10^6$ & $10^6$ & $10^6$\\
    \textbf{batch size} & 256 & 256 & 256 & 1024\\
    \textbf{actor networks} & MLP $(400,300)$ & MLP $(400,300)$ & MLP $(400,300)$ & MLP $(400,300)$\\
    \textbf{critic networks} & MLP $(400,300)$ & MLP $(400,300)$ & MLP $(400,300)$ & MLP $(400,300)$\\
    $\alpha$ & 0.001 & 0.001 & 0.001 & 0.001\\
    $\tau$ & 0.005 & 0.005 & 0.005 & 0.005\\
    $\gamma$ & 0.99 & 0.99 & 0.99 & 0.99\\
    \hline
    \end{tabular}
    \\[10pt]
    
    \caption{
    Table showing the hyperparameters used for each task in our experiments in the parking environment.
    }
    \label{table:hyperparameters1}
\end{table}

\begin{table}
    \textbf{Reacher Environment Hyperparameters}
    \centering 
    \small
    
    \begin{tabular}{|c | c | c | c | c|}
    \hline
     & \textbf{Task 1} & \textbf{Task 2} & \textbf{Task 3}\\
    \hline \hline
    \textbf{episode length} & 200 & 200 & 200\\
    \textbf{buffer size} & $10^6$ & $10^6$ & $10^6$\\
    \textbf{batch size} & 256 & 256 & 256\\
    \textbf{actor networks} & MLP $(400,300)$ & MLP $(400,300)$ & MLP $(400,300)$\\
    \textbf{critic networks} & MLP $(400,300)$ & MLP $(400,300)$ & MLP $(400,300)$\\
    $\alpha$ & 0.001 & 0.001 & 0.001\\
    $\tau$ & 0.005 & 0.005 & 0.005\\
    $\gamma$ & 0.99 & 0.99 & 0.99\\
    \hline
    \end{tabular}
    \\[10pt]
    
    \caption{
    Table showing the hyperparameters used for each task in our experiments in the reacher environment.
    }
    \label{table:hyperparameters2}
\end{table}

Tables \ref{table:hyperparameters1} and \ref{table:hyperparameters2} show the hyperparameters selected for the learning process in each environment and task. Each parameter is described as follows:

\begin{itemize}
    \item ``MLP $(400,300)$'' denotes a multilayer perceptron with two hidden layers consisting of 400 and 300 units, respectively.
    \item $\alpha$ indicates the learning rate used by the ADAM optimizer for gradient updates.
    \item $\tau$ represents the soft update coefficient used in the Polyak averaging of the networks with respect to their corresponding target networks.
    \item $\gamma$ is the discount factor.
\end{itemize}

Most experiments adopt the default hyperparameters provided by the DDPG implementation in \textit{Stable Baselines 3} \citep{stable-baselines3}. The only exception is Task 4 in the parking environment, for which the batch size was increased to allow the algorithms to learn a meaningful policy.

\section{Computing Infrastructure}
All the experiments were implemented in \texttt{Python3.10} and executed on a machine with the following hardware configuration: 
\begin{itemize}
    \item \textbf{CPU}: AMD Ryzen 9 7950X
    \item \textbf{GPU}: NVIDIA RTX 4080 Super
    \item \textbf{RAM}: 32GB DDR5 6000MHz
\end{itemize}

\end{document}